\documentclass[11pt,a4paper]{article}
\usepackage{authblk}
\usepackage[hyperref]{naaclhlt2019}
\usepackage{times}
\usepackage{latexsym}
\usepackage{amssymb}
\usepackage{amsmath}
\usepackage{amssymb}
\usepackage{graphicx}
\usepackage{url}

\aclfinalcopy 

\title{Investigating Machine Learning Methods for \\ Language and Dialect Identification of Cuneiform Texts}

\author[ ]{Ehsan Doostmohammadi}
\author[ ]{Minoo Nassajian}
\affil[ ]{Computational Lingusitics Group,}
\affil[ ]{Sharif University of Technology, Tehran, Iran}
\affil[ ]{\tt{\{e.doostm72, m.nassajian2016\}@student.sharif.edu}}

\date{}

\begin{document}
\maketitle
\begin{abstract}
Identification of the languages written using cuneiform symbols is a difficult task due to the lack of resources and the problem of tokenization.
The Cuneiform Language Identification task in VarDial 2019 addresses the problem of identifying seven languages and dialects written in cuneiform; Sumerian and six dialects of Akkadian language: Old Babylonian, Middle Babylonian Peripheral, Standard Babylonian, Neo-Babylonian, Late Babylonian, and Neo-Assyrian.
This paper describes the approaches taken by 
\tt{SharifCL} \normalfont
team to this problem in VarDial 2019. The best result belongs to an ensemble of Support Vector Machines and a naive Bayes classifier, both working on character-level features, with macro-averaged F\textsubscript{1}-score of 72.10\%.
\end{abstract}

\section{Introduction}
\label{intro}
A wide range of Natural Language Processing (NLP) tasks, such as Machine Translation (MT), speech recognition, information retrieval, data mining, and creating text resources for low-resource languages benefit from
the upstream task of language identification. The Cuneiform Language Identification (CLI) task in VarDial 2019 \cite{vardial2019report} tries to address the problem of identifying languages and dialects of the texts written in cuneiform symbols. 

Identifying languages and dialects of the cuneiform texts is a difficult task, since such languages lack resources and also there is the problem of tokenization. Although there are some work addressing the problem
of tokenization in some of these languages or dialects, there is not any universal method or tool available for tokenization of cuneiform texts, as such a task depends on the rules of that language, simply because cuneiform writing system is a syllabic as well as a logographic one. 
As a result, all the endeavors in this paper are based on character-level features. This work investigates different machine learning methods which are proven to be effective in text classification and compares them by their obtained F\textsubscript{1}-score, accuracy, and training time.

In this paper, we first review the literature of language identification and the work on languages written using cuneiform writing system in \ref{sec:rw}, introduce the models used to tackle the problem of identifying such languages and dialects in \ref{sec:method}, describe the training data in \ref{sec:data}, and discuss the results in \ref{sec:results}.



\section{Related Work}
\label{sec:rw}

The majority of research conducted in the field of language identification has been on textual data. However, there are some studies focusing on speech samples, such as \cite{hategan2009language,ali2015automatic,malmasi2016arabic}.
Language identification systems are meant to distinguish between similar languages \cite{goutte2016discriminating,williams2017twitter}, language varieties \cite{rangel2016low,castro2017smoothed}, or a set of different dialects of the same language \cite{malmasi2016discriminating,el2018arabic}. 
There has also been the annually held VarDial workshop since 2014, which deals with computational methods and language resources for closely related languages, language varieties, and dialects \cite{vardial2017report,vardial2018report}.

Various kinds of features are used to train these systems, including bytes and encodings \cite{singh2007identification,brown2012finding}, characters \cite{van2017exploring,samih2017dialectal}, 
morphemes \cite{gomez2017discriminating,barbaresi2016unsupervised}, and words \cite{duvenhage2017improved,clematide2017cluzh}.

The most recent studies use different language identification methods, such as 
decision trees \cite{bora2018automatic}, 
Bayesian network classifiers \cite{rangel2016low}, 
similarity measures (such as the out-of-place method \cite{jauhiainen2017evaluation}, local ranked distance \cite{franco2017bridging}, and cross entropy \cite{hanani2017identifying}), SVM \cite{alrifai2017arabic}, and neural networks \cite{chang2014recurrent,cazamias2015large,jurgens2017incorporating,kocmi2017lanidenn}. 

To the extent of our knowledge, there is no work addressing the problem of language and dialect identification of cuneiform texts.
Such languages, Sumerian and Akkadian for instance, are considered low-resource languages, meaning that there are only a few electronic resources for cuneiform processing. 
Some of these datasets include \cite{yamauchi2018building} which developed a handwritten cuneiform character imageset,
and \cite{wordflow2018Sumerian} which is an annotated cuneiform corpus with morphological, syntactic, and semantic tags.
Furthermore, there are some early studies on rule-based morphological analyzers for these languages like \cite{kataja1988finite,barthelemy1998morphological,macks2002parsing,barthelemy2009karamel}, and \cite{tablan2006creating}. 

Additionally, a small number of cuneiform text processing tasks have been carried out in which the transliterations of cuneiform characters were considered as the base feature. For instance, \cite{luo2015unsupervised} adapted an unsupervised algorithm to recognize Sumerian personal names. Having transliterated the cuneiform corpus, they utilized the pre-knowledge and applied limited tags to pre-annotate the corpus. 
As another study, \cite{homburg2016akkadian} conducted the first research on word segmentation on Akkadian cuneiform. They used three types of word segmentations algorithms including rule-based algorithms (such as bigram and prefix/suffix), dictionary-based algorithms (like MaxMatch, MaxMatchCombined, LCUMatching, MinWCMatch), and statistical and/or machine learning algorithms (such as C4.5, CRF, HMM, $k$-means, $k$ Nearest Neighbors, MaxEnt, naive Bayes, multi-layer perceptron, and Support Vector Machines (SVM)) which work based on transliterations of cuneiform characters. The paper reports that the dictionary-based approaches obtained the best results. 
In addition, as one of the most recent studies on languages written in cuneiform, \cite{chiarcos2017machine} worked on a machine translation task. The used data consists of unannotated raw transliterations of Sumerian texts with their English translations. They use a morphological analyzer to extract word information to be used in the machine translation task. Moreover, a distantly supervised Part of Speech tagger and a dependency parser are applied to annotate data to facilitate the machine translation task.




\section{Methodology}
\label{sec:method}
We investigated different machine learning methods, all of them based on character-level features, to tackle the problem. The following methods take 1- to 3-gram character TFIDF and 1- to 4-gram character count as input features and were implemented using Scikit-learn \cite{Pedregosa:2011:SML:1953048.2078195}:

\begin{itemize}
\item \bf SVM\normalfont: an SVM with a learning rate of $1\mathrm{e}{-6}$, hinge loss, and \tt elasticnet \normalfont penalty, trained for 5 epochs with a random state of 11.

\item \bf Naive Bayes\normalfont: a multinomial naive Bayes classifier with alpha of $0.14$ and \tt fit\_prior \normalfont as \tt True\normalfont.

\item \bf Ensemble of SVM and naive Bayes\normalfont:
a soft voting classifier which predicts the class label based on the argmax of the sums of the predicted probabilities of the SVM and the naive Bayes models.

\item \bf Random Forest\normalfont: a random forest classifier with 25 estimators of depth 300.
\item \bf Logistic Regression\normalfont: a logistic regression classifier with \tt lbfgs \normalfont optimizer, trained for 100 epochs.
\end{itemize}

We also experimented with deep learning approaches. The following two methods take character embeddings of size 32 for the 256 most common characters as input, and are trained using an Adam optimizer \cite{kingma2014adam} with batch size of 64 and learning rate of $1\mathrm{e}{-4}$:

\begin{itemize}
\item \bf Convolutional Neural Network\normalfont:
The concatenation of the output a set of parallel Convolutional Neural Network (CNN) layers, each with 32 filters and kernel size and stride of 2, 3, 4 and 5 which is fed to a
a dense layer that maps to an $\mathbb{R}^{128}$ space and another one that maps to the $\mathbb{R}^{7}$ space of the labels. We also applied dropout with $0.5$ keeping  rate  on CNN’s  output  and  another  one with the same keeping  rate  on the first  dense  layer's output.

\item \bf Recurrent Neural Network\normalfont:
A Recurrent Neural Network (RNN) with Long Short-Term Memory (LSTM) \cite{hochreiter1997long} cell of size 256 and a dense layer mapping to an $\mathbb{R}^{128}$ space and another one mapping to the $\mathbb{R}^{7}$ space of the labels. We also applied dropout with $0.4$ keeping rate on RNN's output and another one with $0.5$ keeping rate on the first dense layer's output.

\end{itemize}

\section{Data Description}
\label{sec:data}

\begin{figure}[h]
\includegraphics[width=0.46\textwidth]{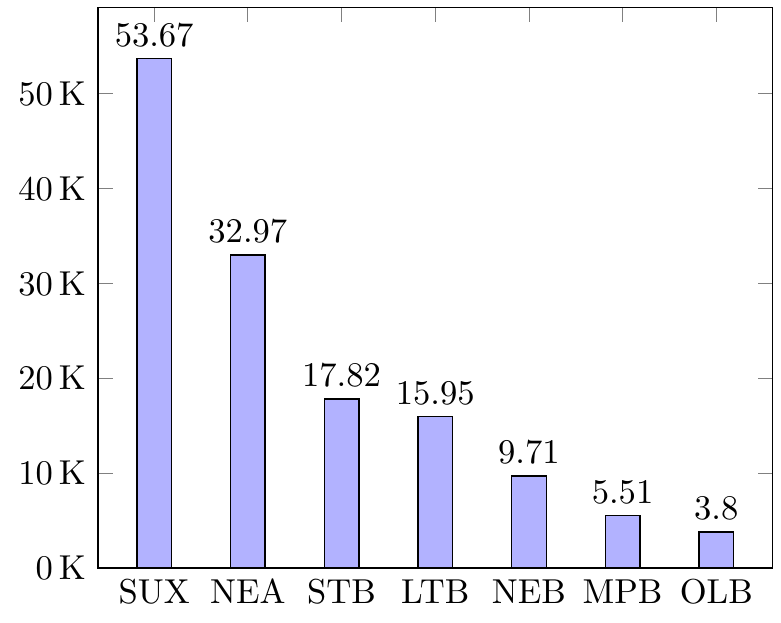}
\caption{Number of samples for each label in the training set (in thousands).}
\label{fig:data}
\end{figure}

The data of CLI shared task is described in \cite{CLIVarDial}.
This data consists of 7 classes: Sumerian (\tt SUX\normalfont), Old Babylonian (\tt OLB\normalfont), Middle Babylonian peripheral (\tt MPB\normalfont), Standard Babylonian (\tt STB\normalfont), Neo-Babylonian (\tt NEB\normalfont), Late Babylonian (\tt LTB\normalfont), and Neo-Assyrian (\tt NEA\normalfont). Figure \ref{fig:data} shows the number of samples for each label in the training data. The whole training data consists of 139,421 samples. The development set comprises 668 and the test set 985 samples per label.

\begin{table}[h]
\center
\begin{tabular}{|c|cc|}
\hline
\bf Label & \bf \# of samples & \bf \% of all\\
\hline
\hline
\tt SUX & 53,673 & 38.49\% \\
\tt NEA & 32,966 & 23.64\% \\
\tt STB & 17,817 & 12.78\% \\
\tt LTB & 15,947 & 11.44\% \\
\tt NEB & 9,707 & 6.96\% \\
\tt MPB & 5,508 & 3.95\% \\
\tt OLB & 3,803 & 2.72\% \\
\hline
\end{tabular}
\caption{Number of samples in the training set for each label and their percentage of a total of 139,421 samples ordered from the highest to the lowest.}
\label{tab:data}
\end{table}

Figure \ref{fig:data} shows that most of the training data belongs to 
\tt{SUX} \normalfont and \tt{NEA} \normalfont
classes. Table \ref{tab:data} contains more detailed information on the data which shows that 86.35\% of the data belongs to four classes of \tt SUX\normalfont, \tt NEA\normalfont, \tt STB\normalfont, \tt LTB\normalfont, whereas only 13.65\% belongs to the other three.

\section{Results and Discussion}
\label{sec:results}
Firstly, we trained the methods described in \ref{sec:method} and evaluated the models on development set. We continued with the best two methods, SVM and NB, and evaluated them on the test set. Table \ref{tab:results-dev} shows the macro-averaged F\textsubscript{1}-score, accuracy, and training time (in seconds) of the five non-deep and two deep methods on the development set. The non-deep models are trained using an \tt Intel(R) Core(TM) i7-7700K CPU @ 4.20GHz \normalfont CPU with 8 threads, and the deep ones using an \tt NVIDIA GeForce GTX 1080 Ti\normalfont.

\begin{table}[h]
\center
\begin{tabular}{|c|ccc|}
\hline
\bf Method & \bf F\textsubscript{1}-score & \bf Accuracy & \bf T. Time \\
\hline
\hline
RF & 0.5201 & 0.5615 & 264.14 \\
LR & 0.6861 & 0.6982 & 40.54 \\
NB & \emph{0.7194} & \emph{0.7301} & \bf 0.15 \\
SVM & \underline{0.7222} & \underline{0.7309} & \underline{1.67} \\
Ens. & \bf 0.7268 & \bf 0.7356 & \emph{3.34} \\
\hline
CNN & 0.6192 & 0.6249 & +4K \\
RNN & 0.6259 & 0.6364 & +4K \\
\hline
\end{tabular}
\caption{Accuracy and F\textsubscript{1}-score on the development set, and the training time (in seconds) of the methods described in section \ref{sec:method}: Random Forest (RF), Logistic Regression (LR), naive Bayes (NB), Support Vector Machine (SVM), Ensemble of the last two (Ens.), and Convolutional and Recurrent Neural Networks (CNN and RNN, respectively). The best result in each column is in bold, the second best underlined, and the third best in italics.}
\label{tab:results-dev}
\end{table}

The ensemble method obtained the best F\textsubscript{1}-score and a very short training time. On the other hand, random forest model suffers from low performance (as it is usually the case in NLP) and a relatively long training time.
The CNN and RNN with embedded characters as input features performed poorly, as it is usually the case in the language identification task \cite{jauhiainen2018automatic}. 
Deep methods see benefit from large amounts of data, however when being trained with fewer data, hyperparameters play a more important role in the results, therefore further tuning them might improve the results in table \ref{tab:results-dev}. As of training time, the naive Bayes method was the fastest and the RNN and the CNN the slowest methods. We also experimented with one-hot encoded characters as RNN's and CNN's input features, which was not fruitful, and therefore are not included in the results.

Table \ref{tab:results-test} shows the results of the SVM and the ensemble of SVM and NB on the test set. The ensemble outperforms SVM, as on the development set.

\begin{table}[h]
\center
\begin{tabular}{|c|cc|}
\hline
\bf System & \bf F1 (macro) & \bf Accuracy \\ 
\hline
\hline
SVM (T) & 0.6660 & 0.6722 \\
SVM (TD) & 0.7171 & 0.7179 \\
SVM + NB (TD) & \bf{0.7210} & \bf{0.7239} \\
\hline
\end{tabular}
\caption{Results of the CLI task on the test set. T stands for training and D for development data. TD means that the model was trained on the combination of training and development data and T, only on the training data. The best result in each column is in bold.}
\label{tab:results-test}
\end{table}

Table \ref{tab:results-best} contains more detailed results of the best performing model on the test set, i.e. the ensemble. It shows the precision, recall, and F\textsubscript{1}-score of the model on each class and their average. The results are ordered based on the F\textsubscript{1}-score. 

\begin{table}[h]
\center
\begin{tabular}{|c|ccc|}
\hline
\bf Label & \bf Precision & \bf Recall & \bf F\textsubscript{1}-score \\
\hline
\hline
\tt LTB & 0.8913 & 0.9655 & 0.9269 \\
\tt MPB & 0.8109 & 0.8579 & 0.8337 \\
\tt OLB & 0.8358 & 0.6924 & 0.7574 \\   
\tt SUX & 0.8273 & 0.6274 & 0.7136 \\
\tt NEA & 0.5621 & 0.8772 & 0.6852 \\  
\tt NEB & 0.6775 & 0.5523 & 0.6085 \\
\tt STB & 0.5515 & 0.4944 & 0.5214 \\ 
\hline
Macro Avg. & 0.7366 & 0.7239 & 0.7210 \\
\hline
\end{tabular}
\caption{Precision, Recall and F\textsubscript{1}-score of all the classes and their macro average ordered from the highest to the lowest F\textsubscript{1}-score.}
\label{tab:results-best}
\end{table}

\begin{figure}
\includegraphics[width=0.49\textwidth]{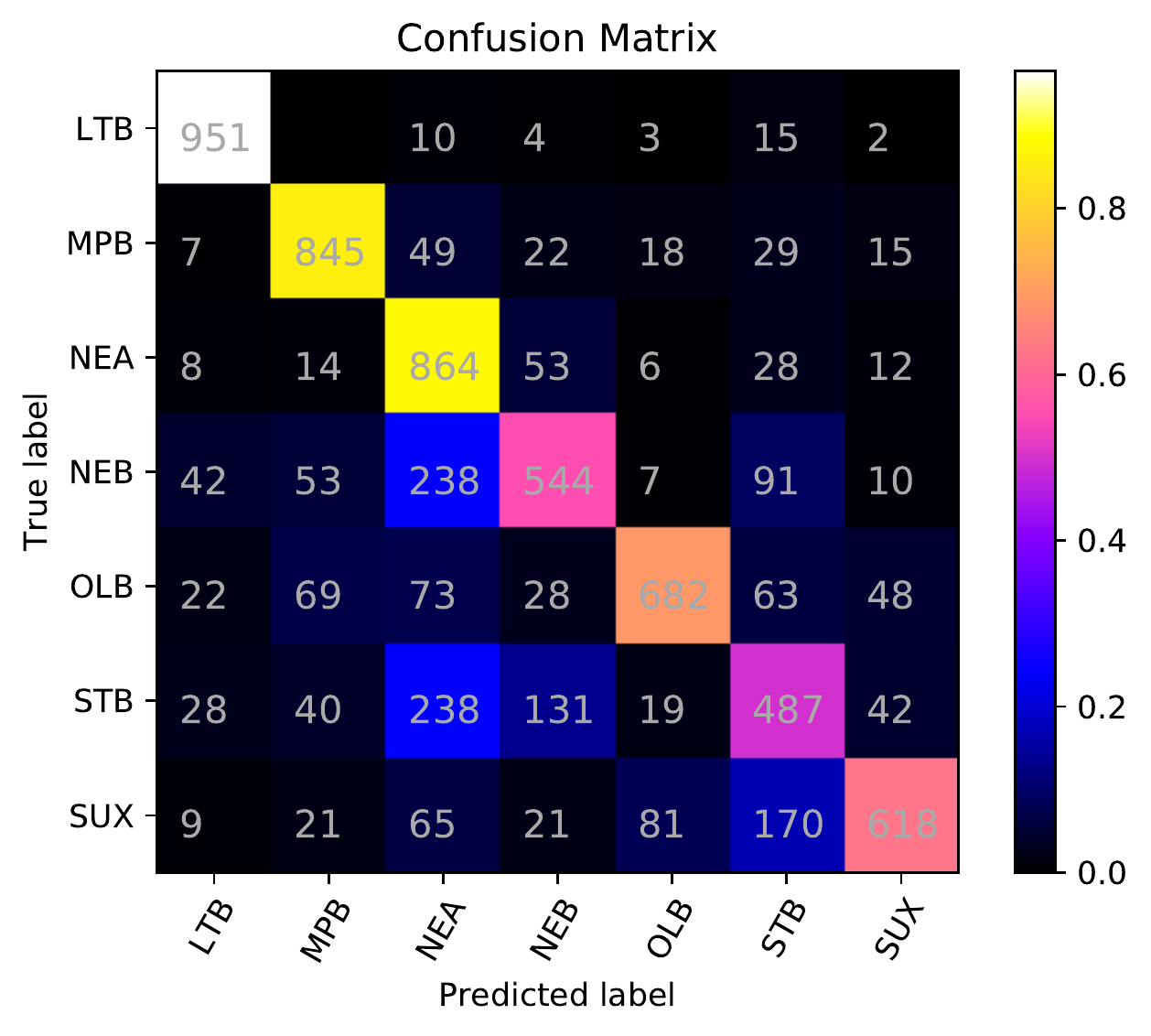}
\caption{Confusion matrix of the ensemble model's results on the test data.}
\label{fig:matrix}
\end{figure}

Considering the results in table \ref{tab:results-best} and the confusion matrix, Late Babylonian (\tt LTB\normalfont) was the easiest class to identify with a recall of 96.55\% and Middle Babylonian Peripheral (\tt MPB\normalfont) the second easiest, with a recall of 85.79\% (with only 5,508 (+668) training samples). Old Babylonian (\tt OLB\normalfont) was also easy to identifiy, especially when we consider its amount training samples, 3,803 (+668).
Standard Babylonian (\tt STB\normalfont) is mainly misclassified as Sumerian, and Neo-Babylonian as Standard Babylonian. Neo-Assyrian (\tt NEA\normalfont) is also among the classes with low F\textsubscript{1}-score, but the model has achieved a very high recall, 87.72\%, in this class. Neo-Assyrian (\tt NEA\normalfont) is mainly misclassified as Neo-Babylonian (\tt NEB\normalfont) and Standard Babylonian (\tt STB\normalfont).

\section{Conclusion}
In this paper, we investigated different machine learning methods, such as SVM and neural networks, and compared their performance in the task of language and dialect identification of cuneiform texts. The best performance was achieved by a combination of SVM and naive Bayes, using only character-level features. It was shown that characters are enough to obtain at least 72.10\% F\textsubscript{1}-score. However, the best model was not able to achieve a good result classifying some of the dialects which indicates a need for other kinds of features, such as word-level ones, and/or embedded or transferred knowledge of these languages and dialects to be used in training the deep models.



\bibliography{vardial}
\bibliographystyle{acl_natbib}

\end{document}